\documentclass[sigconf, superscriptaddress]{acmart}

\AtBeginDocument{%
  \providecommand\BibTeX{{%
    \normalfont B\kern-0.5em{\scshape i\kern-0.25em b}\kern-0.8em\TeX}}}

\setcopyright{acmcopyright}
\copyrightyear{2020}
\acmYear{2020}

\usepackage{xcolor}
\usepackage[shortlabels]{enumitem}
\setlist[enumerate]{nosep}
\usepackage{multirow}
\usepackage{float}
\usepackage{graphicx}
\usepackage{float}
\usepackage{array}
\usepackage{amsmath,lipsum}
\newlength\Origarrayrulewidth
\newcommand{\Cline}[1]{%
  \noalign{\global\setlength\Origarrayrulewidth{\arrayrulewidth}}%
  \noalign{\global\setlength\arrayrulewidth{1.1pt}}\cline{#1}%
  \noalign{\global\setlength\arrayrulewidth{\Origarrayrulewidth}}%
}

\makeatletter
\newcommand\footnoteref[1]{\protected@xdef\@thefnmark{\ref{#1}}\@footnotemark}
\makeatother

\setcopyright{rightsretained}

\begin{document}
\fancyhead{}
\copyrightyear{2020}  
\acmYear{2020}  
\acmConference[CIKM '20]{Proceedings of the 29th ACM International Conference on
Information and Knowledge Management}{October   19--23, 2020}{Virtual Event,
Ireland}
\acmBooktitle{Proceedings of the 29th ACM International Conference on Information
and Knowledge Management (CIKM '20), October   19--23, 2020, Virtual Event,
Ireland}\acmDOI{10.1145/3340531.3412159}
\acmISBN{978-1-4503-6859-9/20/10}

\title{Evaluating the Impact of Knowledge Graph Context on Entity Disambiguation Models}


\author{Isaiah Onando Mulang'}
\email{isaiah.mulang.onando@iais.fraunhofer.de}
\affiliation{%
  \institution{Fraunhofer IAIS and Zerotha Research, Germany}
}

\author{Kuldeep Singh}
\email{kuldeep.singh1@cerence.com}
\affiliation{%
  \institution{Cerence GmbH and Zerotha Research Germany}
}

\author{Chaitali Prabhu}
\email{s6chprab@uni-bonn.de}
\affiliation{%
  \institution{University of Bonn }
  \country{Germany}
}

\author{Abhishek Nadgeri}
\email{abhishek22596@gmail.com }
\affiliation{%
  \institution{Zerotha Research, India}
}

\author{Johannes Hoffart}
\email{johannes.hoffart@gs.com}
\affiliation{%
   \institution{Goldman Sachs, Germany}
}
  \author{Jens Lehmann}
\email{jens.lehmann@cs.uni-bonn.de}
\affiliation{%
  \institution{University of Bonn, Germany}
}


\renewcommand{\shortauthors}{Mulang', IO. et al.}

\begin{abstract}
Pretrained Transformer models have emerged as state-of-the-art approaches that learn contextual information from text to improve the performance of several NLP tasks. These models, albeit powerful, still require specialized knowledge in specific scenarios. In this paper, we argue that context derived from a knowledge graph (in our case: Wikidata) provides enough signals to inform pretrained transformer models and improve their performance for named entity disambiguation (NED) on Wikidata KG. We further hypothesize that our proposed KG context can be standardized for Wikipedia, and we evaluate the impact of KG context on state-of-the-art NED model for the Wikipedia knowledge base. Our empirical results validate that the proposed KG context can be generalized (for Wikipedia), and providing KG context in transformer architectures considerably outperforms the existing baselines, including the vanilla transformer models.
\end{abstract}




\maketitle

\section{Introduction}

Entity Linking (EL) generally consists of two subtasks namely: surface form extraction (mention detection) and named entity disambiguation (NED). A surface form is a contiguous span of text that refers to a named entity. The NED task aims to link the identified named entity to ground truth entities in a given knowledge base~\cite{kuldeep-frankenstein-why-reinvent}. 
For a long time, researchers focused on NED tasks over semi-structured knowledge repositories such as Wikipedia\footnote{\url{https://www.wikipedia.org/}} or publicly available KGs such as DBpedia~\cite{dbpedia-swj}, Freebase~\cite{DBLP:conf/aaai/BollackerCT07}, and YAGO~\cite{yago}. Wikidata~\cite{DBLP:conf/www/Vrandecic12} has recently attracted the community's attention as a rich source of knowledge, and new approaches have been developed to target NED over Wikidata~\cite{cetoli2019neural}. 



\textbf{Motivation}: A peculiarity of Wikidata is that the contents are collaboratively edited. As at April 2020; Wikidata contains 83,151,903 items and a total of over 1.2B edits since the project launch\footnote{\url{https://www.wikidata.org/wiki/Wikidata:Statistics}}. Considering Wikidata is collaboratively edited, the user-created entities add additional noise and non standard labels ((e.g. labels have several numeric and special, non-alphanumeric ASCII characters, also contains multi word labels up to 62 words, etc)) ~\cite{Mulang2019ContextawareEL} since users do not follow a strict naming convention. For instance, there are 17,88,134 labels in which each label matches with at least two different URIs. Hence, NED on Wikidata is quite challenging as pointed out by initial studies~\cite{Mulang2019ContextawareEL,sakor2019falcon}. For example, consider the sentence from Wikidata-Disamb \cite{cetoli2019neural} dataset: \textit{"the short highway in New South Wales and the Australian Capital Territory in Australia, it is part of Sydney-Canberra \textit{\underline{National Highway}} link"}. The entity surface form \textit{\underline{National Highway}} matches four(4) different entities in Wikidata that share the same entity label (i.e., "National Highway") while 2,055 other entities contain the whole mention in their labels. The correct entity \texttt{wikidata:Q1967298}\footnote{\texttt{wikidata:Q1967298} binds to  \url{https://www.wikidata.org/wiki/Q1967298}} refers to \texttt{Highway System of Australia}, whereas \texttt{wikidata:Q1967342} refers to the \texttt{highway system in India}. Having these two entities as candidates may require extra information in addition to the surface form or the sentence context. Attention-based Neural Networks \cite{Attention-DBLP:journals/corr/VaswaniSPUJGKP17} and pretrained transformer models \cite{XLNet-DBLP:journals/corr/abs-1906-08237,RoBERTa-DBLP:journals/corr/abs-1907-11692}, have provided an avenue for encoding the context within text, howbeit, in cases such as our example, pure textual context may not be sufficient \cite{Liu2019KBERTEL}. As such, a method to obtain matching contextual information from the KG itself could be beneficial to disambiguate in such close scenarios. Inspired by the work of \cite{cetoli2019neural} and inherited Wikidata NED challenges (nonstandard multi-word, long, implicit, case-sensitive), we hypothesize that the performance of pretrained transformer models improves by considering further context from the KG. We investigate three research questions: \textbf{RQ1:}How does applying KG context impact the performance of transformer models on NED over Wikidata? \textbf{RQ2:} What is the performance of different configurations of KG context as new information signals on the NED task? \textbf{RQ3:} Can we generalize our proposed context in a state of the art NED model for other knowledge bases such as Wikipedia? The structure of the paper is follows: next section defines the task followed by related work in \ref{sect:related-work}. Section \ref{sect:approach} describes approach and we present experiments in section \ref{sec:evaluation}. We conclude in section \ref{sec:conclusion}.

\begin{figure*}[h!]
\centering
\includegraphics[width=17cm]{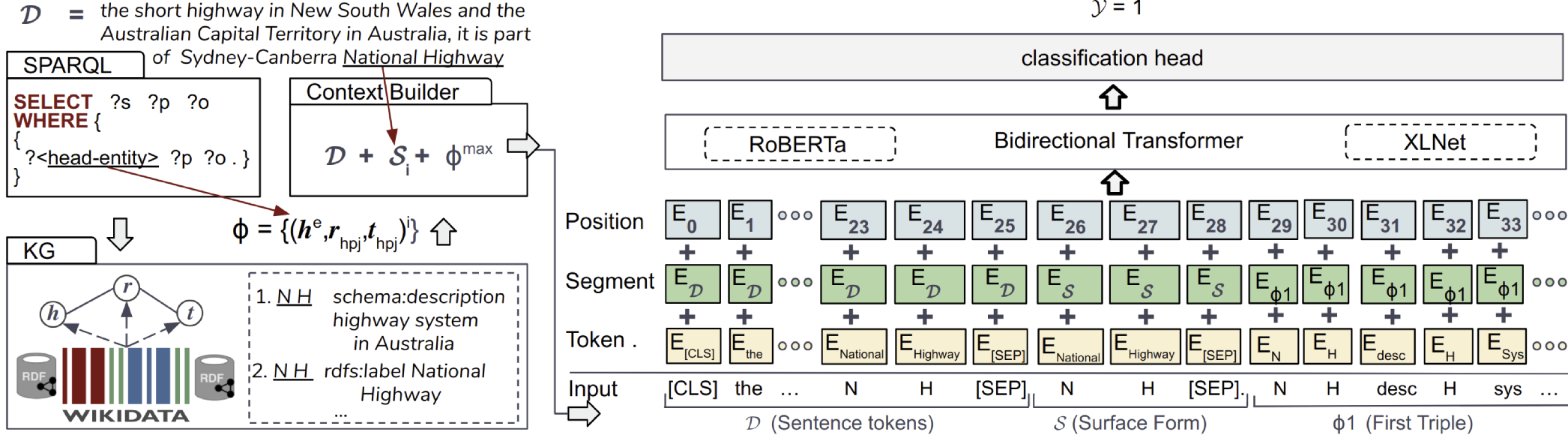}
\vspace{-2mm}
\caption{\textit{Overall Approach : $\Phi$ refers to the ordered set of triples from the KG for a candidate entity while $\Phi^{max} \subseteq \Phi$, is the maximum number of triples that fits in the sequence length. For brevity: N \textrightarrow{} "National", H \textrightarrow{} "Highway", desc \textrightarrow {} "description"}}
\label{fig:approach}
\vspace{-2mm}
\end{figure*}
 
\vspace{-2mm}
\section{Task Definition}\label{sect:problem}
Given a sentence, a recognized entity surface form, a set of candidate entities, and a Knowledge Graph ($ KG $), the objective is to select the entity within the $ KG $ that matches the surface form in the text. A sentence $\mathcal{D} = \{ w_1,w_2,...,w_n\}$ is a set of tokens of length $n$. The set of entity surface forms $S = \{s_1,s_2,...,s_k\} : s_i = (w^\prime\in\mathcal{D})+ $ contains recognized entities s.t. each $s$, spans one or more tokens in $\mathcal{D}$. We view a KG as a labeled directed graph. A $KG = (\mathcal{E},\mathcal{H}^+,\mathcal{R})$ is a 3-tuple where: i) set $\mathcal{E}$ of all entities represent the vertices. ii) set $ \mathcal{R} $ is the set of all edges between the entity instances in the graph. And; iii) $\mathcal{H}^+ \subseteq \mathcal{E} \times \mathcal{R} \times \mathcal{E} $ is the ordered set of all triples.
\noindent
\noindent
The function $ \ell (a\in{\mathcal{E}\cup \mathcal{R}})$ is defined that retrieves labels of any given entity or relation from the KG. 


\textit{Candidate Entities} : A set containing selected entities from the KG given as:   $\mathcal{E}^\prime=\{e^\prime_1,e^\prime_2,...,e^\prime_m\}$ where $e^\prime_j \in \mathcal{E}$ and  $\mathcal{E}^\prime = \sigma (s_i \in S) $, is obtained by a semantic selection operation $\sigma$ on a given surface form. This paper addresses the problem of named entity disambiguation which selects an entity $e^c \in \mathcal{E}^\prime$ that matches the textual mention $ s \in S $ . We view this task as a classification $f= Classify(h(X)) $ on the conditional probability $h(X) = P(Y=1|X)$. Taking $x\in X= (s,e^\prime; \theta)$ we study configurations of the context parameters $\theta$. 
\section{Related Work}
\label{sect:related-work}

There is a wide variety of approaches in the literature for Entity Linking (EL) ranging from graph traversal \cite{UsbeckNRGCAB14}, rule-based \cite{sakor2019old}, to Neural Network-based approaches \cite{kolitsas-etal-2018-end,Mulang2019ContextawareEL}.
For detailed information on entity linking we refer to the surveys in \cite{Survey-shen-et-al,Balog2018}. Herein, we restrict ourselves to very closely related recent literature. Advances in Neural Networks and the introduction of self-attention based techniques~\cite{Attention-DBLP:journals/corr/VaswaniSPUJGKP17} that allow for encoding of contextual meaning from text have advanced research in EL. Work in \cite{DBLP:conf/emnlp/GaneaH17} improved EL performance by proposing a new approach that combines deep learning with more traditional methods such as graphical models and probabilistic mention-entity maps. Research in~\cite{kolitsas-etal-2018-end} aims to achieve an end-to-end EL through context-aware mention embeddings, entity embeddings, and a probabilistic mentions.

In the meantime, research on the learning of contextual data has advanced in two directions. On one hand, the powerful pretrained transformer models~\cite{RoBERTa-DBLP:journals/corr/abs-1907-11692,XLNet-DBLP:journals/corr/abs-1906-08237} have emerged as state-of-the-art for representing context within text and have seen burgeoning reuse through fine-tuning for several NLP tasks \cite{DBLP:journals/corr/abs-2001-01447}. On the other hand, KGs are increasingly being seen as a source of additional knowledge for Neural Networks. For instance researchers in~\cite{BigGraph-DBLP:journals/corr/abs-1903-12287} recently released an embedding library for the Wikidata KG while the work by~\cite{Liu2019KBERTEL} introduced an extension of BERT (KBERT) in which KG triples are injected into the sentences as domain knowledge. Specific to the EL task, the work by~\cite{Mulang2019ContextawareEL} employs information from a locally derived KG to improve the performance of end-to-end EL using attention-based Neural Networks. Researchers in~\cite{cetoli2019neural} fetched a significant amount (as high as 1500) of 2-hop KG triples and used Recurrent Neural Networks (RNN) to encode this information. In this paper, we argue that a slight amount of KG triple context is enough for a pretrained transformer. We study the different configuration of KG context on transformer models to target NED on Wikidata.
\section{Approach} \label{sect:approach}
Figure ~\ref{fig:approach} illustrates the overall approach. For the classification :
$ f(h(s,e^\prime; \theta)) = y $ such that $ s $, the mentioned surface form, and $e^\prime$, the candidate entity, are known. A set of contextual parameters $ \theta $ is then provided to the model. By adding the original sentence as part of the input, we let the model learn source context. Our approach then models a set of information from the target KG in the form of KG triples $ \Phi $ as context. The aim is to maximize both the true positives and true negatives such that, for every input, if $ y=1 $ then the $ e^c $ is the ground truth entity of $s$ in the KG. The classifier employs the binary cross-entropy loss.

\begin{figure}[!h]
\includegraphics[width=8.5cm]{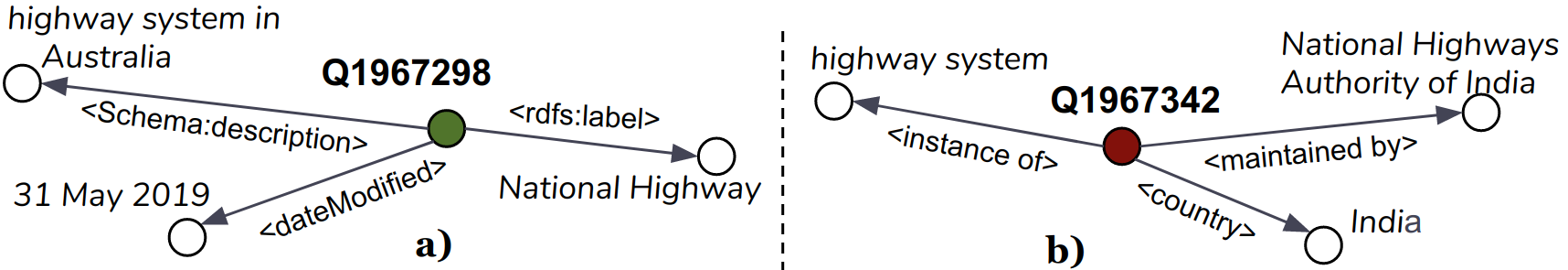}
\vspace{-2mm}
\caption{\textit{\small{KG context : Top three 1-hop triples from Wikidata for the two entities with same label: National Highway.}}}
\label{fig:context} 
\vspace{-3mm}
\end{figure}

\textbf{Knowledge Graph Context: }We use a SPARQL endpoint to fetch triples of the identified entity in the sentence. There are two sets of triple configurations considered in our experiments, depending on the hop counts from the head entity. The parameter $\Phi$ is therefore an ordered set of of triples $( h^e,r_{hp},t_{hp})^i $ such that $h^e$, the head (subject) of any triple is the candidate entity to be classified whereas $hp = 1|2$ is the hop count. The $i$ refers to the position of the triple in the set and can range between 1 and over 1000. To formulate our input, we consider the natural language labels of the retrieved triples $l_{h^e}$ , $l_{r}$ , $l_{t}$. A triple is therefore verbalized into it's natural language form: "$l_{h^e} [white space] l_{r} [white space] l_{t}$". The sequence of these verbalized triples are appended to the original sentence and surface form delimited by the [SEP] token. Figure \ref{fig:approach} shows how the context input is handled such that the Segment Embeddings for every triple is different and provides a unique signal to the tokens at the embeddings layer of the network.  When the total number of triples is too many, we use the maximum sequence length to limit the input where the final context representation $\Phi^{max} \subseteq \Phi $. The values of $\Phi^{max}$,for entity: Q1967298 in figure \ref{fig:context}, is given as: 
\texttt{\footnotesize{[National Highway description highway system in Australia [SEP] 
National Highway label National Highway [SEP] National Highway date modified 31 May 2019 [SEP]]}}
\section{Evaluation}
\label{sec:evaluation}
\textbf{Datasets:} The first dataset is \textit{Wikidata-Disamb}\cite{cetoli2019neural}, which aligns Wiki-Disamb30 \cite{TagMe-DBLP:journals/corr/abs-1006-3498} to Wikidata entities, and adds closely matching entities as negative samples to every entity in the dataset. It consists of 200,000 Train and 20,000 Test samples. We also consider the \textit{ISTEX} dataset introduced by \cite{Opentapioca-DBLP:journals/corr/abs-1904-09131}, extracted from scientific publications and contains 1000 author-affiliation strings from research articles aligned to Wikidata. For generalizing the impact of KG context, we choose standard Wikipedia dataset: AIDA-CoNLL \cite{DBLP:conf/emnlp/HoffartYBFPSTTW11}. We  aligned its Wikipedia entities to corresponding Wikidata mentions to fetch the KG triples. Datasets are accompanied with a pre-computed candidate list.\\
\textbf{Baselines: }
We compare our results with three types of baselines. First is \cite{cetoli2019neural}, which experimented with numerous configurations of KG context on Long Short Term Memory (LSTM) networks and reported an ablation of these configurations. These models were augmented with a massive amount of 1\&2-hop KG triples. We also run the model on the ISTEX dataset to enable performance comparison. We create a second set of baselines by employing the vanilla transformer models of RoBERTa and XLNet(i.e., transformers without KG context) on Wikidata-Disamb and ISTEX. We fine-tuned vanilla models on Wikidata-Disamb training set. 
For AIDA-CoNLL, we chose \cite{DBLP:conf/emnlp/YangGLTZWCHR19} as our underlying model which is the second \textbf{peer reviewed} SOTA on this dataset. Authors used Wikipedia descriptions as a context for candidate entities, and we replaced this context with our proposed 1-hop KG triple context fetched from Wikidata triples of corresponding Wikipedia entities. We verbalized the fetched triples, as described in our approach. 
\vspace{-2mm}

\textbf{Model Parameters: } We chose two state of the art transformer architectures: RoBERTa \cite{RoBERTa-DBLP:journals/corr/abs-1907-11692}, and XLNet  \cite{XLNet-DBLP:journals/corr/abs-1906-08237} and fine-tune them using Wikidata-Disamb30 training set. We report P,R,F values following the baseline of Wikidata-Disamb and ISTEX dataset. For each vanilla Transformer architecture, we add a classification head. The maximum sequence length for the inputs in both models is fixed at 512 tokens, and we use this to limit the amount of KG context to feed. We publicly release code, datasets, training details, and results for reusability and reproducibility. On AIDA-CoNLL, We use open source implementation of \cite{DBLP:conf/emnlp/YangGLTZWCHR19} for feeding the KG context and report In-KB accuracy as prior work(s).

\begin{table}[ht!]
    \centering
    \small
    \begin{tabular}{p{4cm}|p{1cm}|p{1cm}|p{1cm}}
     \Cline{1-4}
     Model & Prec & Recall & F1\\
    \cline{1-4}
     LSTM + RNN-triplets \cite{cetoli2019neural}   &90.1 &92.0 &91.1\\
     LSTM+RNN-triplets+ATN\cite{cetoli2019neural} & 90.2  & 93.0  & 91.6\\
     \cline{1-4} 
     RoBERTa - without KG context  & \textcolor{gray}{\textbf{89.09}} & \textcolor{gray}{\textbf{84.67}} & \textcolor{gray}{\textbf{86.23}}\\
     XLNet-without KG context & 89.32 & 87.62 & 88.46\\
     \cline{1-4}
     \textit{Our Contextual models }& & & \\
     RoBERTa + 1-hop KG Context & 91.48 & \textbf{93.23} & \textbf{92.35}
    \\
      RoBERTa + 1\&2-hop KG Context  & 89.88 & 87.64 & 88.75
    \\
     XLNet + 1-hop KG Context & 91.55 & 93.14 & 92.34\\
     XLNet + 1\&2-hop KG Context  & \textbf{91.93} & 92.36 & 92.14
    \\
     \Cline{1-4}
     
    \end{tabular}
\caption{\small{\textit{Comparison of our model against baselines on the \textit{Wikidata-Disamb} dataset. Best results in dark bold}}}
\label{tab:results_baseline_1}
\vspace{-2mm}
\end{table}

\textbf{Results and Discussion}: Table \ref{tab:results_baseline_1} shows the results from evaluating our approach against the baselines on the Wikidata-Disamb30 and table \ref{tab:results_baseline_lm_2} indicates the performance of the models on the ISTEX dataset. The results in table \ref{tab:results_baseline_lm_2} obtained by running the same model trained on the Wikidata-Disamb30 dataset (also for baseline) with context, but during testing, no more context is provided. Based on our results, we postulate that although the Transformer based language models are trained on huge corpus and possess context for the data, \textit{they show limited performance even against the RNN model. This RNN model \cite{cetoli2019neural} uses GloVe embeddings} together with task-specific context (cf. Table \ref{tab:results_baseline_1}).  However, the transformer models outperform the baseline models when fed with our proposed KG context. For instance, (cf. Table \ref{tab:results_baseline_1}), RoBERTa, with a 1-hop context, can correctly link entities in additional 1127 sample sentences in the test set (flipped from incorrect to the correct predictions) compared to its vanilla setting. These samples have 997 unique Wikidata IDs. 
\begin{table}[!htp]
    \centering
    \small
    \begin{tabular}{p{5cm}|p{0.65cm}|p{0.7cm}|p{0.66cm} }
     \Cline{1-4}
     Model & Prec & Recall & F1\\
     \cline{1-4}
      LSTM + RNN of triplets + ATN \textit{\cite{cetoli2019neural}}   & 86.32    &\textbf{96.38} &90.97\\
     \cline{1-4}
     RoBERTa + 1-hop Triples (\textbf{Ours})&   91.70  & 91.98 & 91.84\\
     XLNet + 1-hop KG Context(\textbf{Ours}) &   \textbf{96.39}  & 89.11 &\textbf{92.61}\\
     \Cline{1-4}
    \end{tabular}
    \caption{\textit{Our models against baseline on ISTEX dataset}}
    \label{tab:results_baseline_lm_2}
    \vspace{-2mm}
\end{table}
\vspace{-4mm}
These results also indicate that the transformer models achieve better precision compared to recall; this is clear in table \ref{tab:results_baseline_lm_2} and interpreted as follows: our model is more likely to classify an entity as the correct entity only when it is true (few false positives).
For brevity, the detailed analysis of each experimental setup and corresponding data can be found in our Github\footnote{\url{https://github.com/mulangonando/Impact-of-KG-Context-on-ED}}.

\begin{table}[ht!]
    \centering
    \small
    \begin{tabular}{p{4.2cm}|p{1.6cm}}
     \Cline{1-2}
     \textbf{Model} & In-KB. Acc.\\
    \cline{1-2}
     Yamada et al. (2016) \cite{DBLP:conf/conll/YamadaS0T16}   &91.5\\
     Ganea\&Hofmann (2017) \cite{DBLP:conf/emnlp/GaneaH17} & 92.22$\pm$0.14 \\
     Yang et al. (2018) \cite{DBLP:conf/naacl/YangIR18} & 93.0\\
     Le\&Titov (2018) \cite{DBLP:conf/acl/LeT19}& 93.07$\pm$0.27 \\
     DeepType (2018) \cite{DBLP:conf/aaai/RaimanR18} & \underline{94.88}\\
     Fang et al. (2019) \cite{DBLP:conf/www/FangC0ZZL19} & 94.3 \\
     Le\& Titov (2019) \cite{DBLP:conf/acl/LeT19} & 89.66$\pm$0.16 \\
     DCA-SL (2019)\cite{DBLP:conf/emnlp/YangGLTZWCHR19} & 94.64$\pm$0.2\\
     Chen et al. (2020) \cite{chen2020}  & 93.54$\pm$0.12\\
     \textbf{DCA-SL + Triples(ours)} & \textbf{94.94$\pm$0.2}\\
     \Cline{1-2}
    \end{tabular}
\caption{\textit{\small{Generalizability Study: Comparison of KG Context based model against baselines on the \textit{AIDA-CONLL} dataset}. Best value in bold and previous SOTA value is underline.}}
\label{tab:results_baseline_3}
\vspace{-2mm}
\end{table}

Concerning \textbf{RQ2}, our results indicate that including triples from higher hop counts either exhibit an inverse impact on the performance or have minimal effect on overall model behavior (cf. Table \ref{tab:results_baseline_1} RoBERTa vs. XLNet 2-hop values). This signals that the further away we drift from the head entity, the noisier the signal provided by the context added. As such, we did not extend evaluation to higher triple hops. However, we can observe that XLNet shows a more stable behavior in cases when the excess context is provided as it can preserve already learned information. It is in contrast to RoBERTa, which loses necessary signals in an attempt to learn from the extra context. The amount of data fed as the context in our models is minimal (up to 15 1-hop triples). In contrast,  the best performing model from work in \cite{cetoli2019neural}, was fed up to 1500 1+2-hop triples. Our best performance can then be attributed to the quality of textual context learned by the transformers as well as the optimal choice of KG-triples context.\\
\textbf{Generalizing KG Context:} We induced 1-hop KG context in DCA-SL model \cite{DBLP:conf/emnlp/YangGLTZWCHR19} for candidate entities. The replacement of the unstructured Wikipedia description with structured KG triple context containing entity aliases, entity types, consolidated entity description, etc. has a positive impact on the performance. Our proposed change (DCA-SL + Triples) outperforms the baselines for Wikipedia entity disambiguation(cf. Table \ref{tab:results_baseline_3}). Please note, out of 207,544 total entities of AIDA-CoNLL datasets, 7591 entities have no corresponding Wikidata IDs. Even if we do not feed the KG context for 7591 entities, the performance increases. It validates our third research question (\textbf{RQ3}), and we conclude KG triple context can be standardized for the NED task for Wikipedia. 

\section{Conclusion}
\label{sec:conclusion}
In this paper, we study three closely related research questions. We demonstrate that pretrained Transformer models, although powerful, are limited to capturing context available purely on the texts concerning the original training corpus. We observe that an extra task-specific KG context improved the performance. However, there is a limit to the number of triples as the context that can improve performance. We note that 2-hop triples resulted in negative or little impact on transformer performance.  Our triple context can be generalized (for Wikipedia) and observes a positive effect on the NED model for Wikipedia, leading into a new SOTA for AIDA-CoNLL dataset. For the future work, it would be interesting to understand which triples negatively impact the context and how to select the "optimal choice of KG-triples context," considering we rely on the triple in the same order of the SPARQL endpoint returned results. As a viable next step, we plan to study independent effect of various KG attributes (entity properties such as aliases, descriptions, Instance-of, etc.) on NED models' performance.
\section{Acknowledgments}
This work is co-funded by the Federal Ministry of Education and Research's (BMBF) Software Campus initiative under the Answer-KiNG Project.




\bibliographystyle{ACM-Reference-Format}
\bibliography{bibliography}










\end{document}